\documentclass{article}

\usepackage[nonatbib, final]{neurips_2020}
\usepackage[utf8]{inputenc} 
\usepackage[T1]{fontenc}    
\usepackage{hyperref}       
\usepackage{url}            
\usepackage{booktabs}       
\usepackage{amsfonts}       
\usepackage{nicefrac}       
\usepackage{microtype}      
\usepackage{amsmath}
\usepackage[ruled,vlined]{algorithm2e}
\usepackage{multirow}
\usepackage{array}
\usepackage{booktabs}
\usepackage{graphicx}

\usepackage{color}

\title{Reprogramming Language Models for Molecular Representation Learning}

\author{%
  Ria Vinod\thanks{Work done during internship at IBM Research} \\
  Department of Electrical Engineering and Computer Science\\
  University of California, Berkeley\\
  \texttt{ria.vinod@berkeley.edu} \\
  \And
  Pin-Yu Chen \\
  IBM Research \\
  Yorktown Heights, NY \\
  \texttt{pin-yu.chen@ibm.com} \\
  \AND
  Payel Das \\
  IBM Research \\
  Yorktown Heights, NY \\
  \texttt{daspa@us.ibm.com} \\
}

\begin{document}

\maketitle

\begin{abstract}
Recent advancements in transfer learning have made it a promising approach for domain adaptation via transfer of learned representations. This is especially relevant when alternate tasks have limited samples of well-defined and labeled data, which is common in the molecule data domain. This makes transfer learning an ideal approach to solve molecular learning tasks. While Adversarial Reprogramming has proven to be a successful method to repurpose neural networks for alternate tasks, most works consider source and alternate tasks within the same domain. In this work, we propose a new algorithm, Representation Reprogramming via Dictionary Learning (R2DL), for adversarially reprogramming pretrained language models for molecular learning tasks, motivated by leveraging learned representations in massive state of the art language models. The adversarial program learns a linear transformation between a dense source model input space (language data) and a sparse target model input space (e.g., chemical and biological molecule data) using a k-SVD solver to approximate a sparse representation of the encoded data, via dictionary learning. This method achieves the baseline established by state of the art toxicity prediction models trained on domain-specific data using only a standard well-trained text classifier, thereby establishing avenues for domain-agnostic transfer learning for tasks with molecule data.
\end{abstract}

\section{Introduction}
Deep learning has proven to be an extremely successful tool for various applications for the natural sciences. While works like MoleculeNet have made significant progress in publishing benchmarks on various molecular learning tasks, curating substantial, well-structured and labeled molecule datasets remains a critical constraint in training high performing models from scratch to establish baselines for a multitude of tasks \cite{moleculenet}. A lack of substantive training datsets motivates transfer learning to be a natural approach to solve problems in the molecule data domain, as it has proven to be a successful technique to solve new tasks using learned representations from source model domains \cite{pan_yang}. In this paper, we consider an empirically successful transfer learning method, adversarial reprogramming, where a learned function adversarially perturbs input samples to the model such that the model can perform a task chosen by the adversary \cite{elsayed}. The motivations of the transfer learning approach lie in the recent successes of powerful general language models \cite{bert}, and being able to leverage these learned representations when applied to molecule data that can be treated as sequence data. However, the adversarial reprogramming method is well suited to models operating within a continuous input space (such as images), but discrete input spaces require that we learn a mapping between source-model space and target-model space. Reprogramming of text classifiers and language models has been explored \cite{neekhara}, as well as knowledge injection for general language models \cite{knowledge_injection}, but such approaches do not investigate mappings between domains that require a very high representational capacity (from language data to molecule data). This shift in domain poses a significant challenge through the reduction of dimensionality of the input space. We end up with an overdetermined system, as we have more observations in the source-model input space than in the target-model input space. To address this, we use a dictionary learning approach to encode a sparse representation of the embeddings of the input data to the pre-trained language model (text classifier). Training the adversarial program (AP) via dictionary learning enables us to approximate a linear transformation (mapping) between the input spaces of the source and target models, it learns the optimal coefficients of the atoms in our molecule data to represent the dictionary. When compared to the baselines of training AMP models from scratch, repurposing a sentiment classifier via R2DL outperforms the baseline for AMP/non-AMP (90.01 \% vs 88.0 \%) and achieves approximately the same progress on toxicity prediction (89.34 \% vs 93.7 \%).

\section{Related Work}

\subsection{Adversarial Machine Learning and Reprogramming}
Adversarial attacks on ML models in discrete input spaces have been found to cause models to misclassify input samples by manipulating tokens in sequence data \cite{adv_ml}. Adversarial Reprogramming (AR) is a technique \cite{elsayed} that is able to repurpose a model (without significant changes to the architecture and parameters) by training an adversary to optimally transform input data such that we can choose the output of the model. This method has been proven to be successful in both, white-box and black-box settings \cite{bar}. The depth and the size of large general language models make them ideal candidates to be adversarially reprogrammed, as modifying the internal architecture or finetuning over 1 billion parameters is infeasible for a traditional transfer learning approach. The work in \cite{neekhara} assumes an exposed model, where they use the learned parameters of the pre-trained source model to generate context-based vocabulary map. However, the hypothesis of comparable vocabulary size does not hold when the source data are English vocabularies (on the order of $\sim$ 10 million) and the target data are amino acids (on the order of $\sim$ 20). To that end, we introduce dictionary learning to train our AP.

\subsection{Dictionary Learning}
Typical amino acid/nucleotide representations of biological molecules or widely used SMILES representation of chemical molecules have fewer distinct tokens than that of language data. The significant reduction in dimensionality of the embedded space of English data to molecule data makes finding a mapping non-trivial. Results in \cite{bengio_rep} demonstrate that that representation learning algorithms have an advantage in transfer learning methods as they capture features relevant to alternate tasks. We require a mapping with high representational capacity. Work in \cite{goodfellow_inv_networks} demonstrates that a sparse encoding is distributed and highly expressive: we can represent $O(2^k)$ input regions with only $O(N)$ parameters. Distributed representations can be clustered to extract relevant features where component extraction algorithms can find the optimal representation (a dictionary). We use a k-SVD approximation algorithm \cite{k-SVD} to mitigate computational expenses. 

\section{Representation Reprogramming via Dictionary Learning (R2DL): Algorithm and Method}

\subsection{Problem Formulation}
We are given a pretrained classifier, $X$, a source-task dataset $\{V_S\}^{n}_{i=1}$ and target-task dataset $\{V_T\}^{m}_{j=1}$. The embedded matrices are $V_T$ and $V_S$ respectively. We can encode an output label mapping function $h : l_{V_S} \longmapsto l_{V_T}$. We then aim to train an adversarial program (AP) $f_\theta$ that finds the optimal coefficients $\theta$ of our atoms in $V_T$ to represent a sparse encoding of the dictionary, $V_S$ such that $V_T = V_S \theta$. The AP $f_\theta$ is used to perform the target task through the transformation $h(X(f_\theta(t_i))$ where $t_i$ is a molecule data sample. While we do not make any modification the parameters or architecture of $X$, we assume access to the gradient $\nabla_{\theta} X$ for loss evaluation during training. 

\subsection{Adversarial Program}
To reprogram the pretrained classifier, we use a similar structure as introduced in \cite{neekhara}, $f_\theta : t_i \longmapsto s_i$ where $t_i \in V_T$ and $s_i \in V_S$. Dimension of the input space of $V_S$ and $V_T$ is $|V_S|$, and $|V_T|$ respectively, where $|V_T| \ll |V_S|$. The AP $f_\theta$ is parametrized by $\theta \in \mathbf{R}^{|V_S| \times |V_T|}$, which represents the coefficients of the atoms in $V_T$ such that $V_T = V_S \theta$ and $V_S$ is our dictionary. The observation of $|V_T| \ll |V_S|$ requires that our mapping has high representational capacity, so we encode a sparse representation of $V_S$, to extract relevant features from the embeddings of the source-model vocabulary $\{V_S\}^{n}_{i=1}$ for the alternate task. To that end, approximate the dictionary, we use a k-SVD solver to optimize over the cross entropy loss for updates to $\theta$.

\subsection{k-SVD}
We define $\theta \in \mathbf{R}^{|V_S| \times |V_T|}$ and use $\theta_t$ to denote its $t$-th column, where a signal (embedding vector) $v_t$, can be represented as a sparse linear combination of the signal atoms of columns of $V_S$, $v_t = V_S \theta_t$. $v_t$ is the representation of the AMP input sample in the dictionary space and satisfies $||v_t - V_s \theta_t||_p \leq \epsilon$. An exact solution $\theta_t^*$ such that $v_t =  V_S \theta_t^*$ is computationally expensive to find, and is subject to various convergence traps, so for the purpose of our efficient AP approach we approximate $v_t \approx V_s \theta_t $ by limiting the number of iterations to converge to a solution for $v_t$. Our optimization problem for finding the optimal sparse representation of input data samples in $V_T$ is then minimize $||\theta||_0$ subject to $||v_t - V_s \theta_t ||_1 \leq \epsilon$ to enforce a sparse solution \cite{k-SVD},
where $||\cdot||_0$ denotes the $l_0$ norm. While algorithms exist to choose an optimal dictionary (an exact solution to k-SVD) that can be continually updated \cite{k-SVD}, we penalize computational expense over performance for the purpose of maintaining an efficient solution (at the cost of statistically insignificant improvements in accuracy) by using a predetermined number of steps for an approximate solution, $V_S$ that encodes the atoms of $V_T$. $\theta$ is then used to evaluate the cross entropy loss on $h(X(f_\theta(V_T))$, which will be updated in the AP $f_\theta$.


\section{Experiments}

\SetKwInput{KwInput}{Input}                
\SetKwInput{KwOutput}{Output}              

\begin{algorithm}[H]
\SetAlgoLined
\KwInput{Pretrained sentiment classifier $X$, source model training data $\{V_S\}^{n}_{i=1}$, target model training data $\{V_T\}^{m}_{j=1}$, maximum number of iterations $T_1$ for updates to $\theta$, number of iterations $T_2$ for k-SVD, output-label mapping function $h$(·), step size $\{\alpha_i\}^{T_1}_{i=1}$}
\KwOutput{Optimal adversarial program parameters $\theta$}
 1. Random initialization of $\theta$\;
 2. Define objective function for k-SVD from (1) \;
 3. \While{$t_1 \leq T_1$}{
      4. \While{$t_2 \leq T_2$}{ 
        use approximate k-SVD to solve $V_T \approx  V_S \theta$ \; 
        $t_2 \longleftarrow t_2 + 1$ \;
    }
  \textbf{\#Loss evaluation for $\theta$} \;  
   $\theta \longleftarrow \theta - \alpha_{t_1} \cdot \nabla_\theta h(X(f_\theta(V_T)) $\;
  
  $t_1 \longleftarrow t_1 + 1$ \;
 }
 
 \caption{Representation Reprogramming via Dictionary Learning (R2DL)}
\end{algorithm}

\subsection{Baseline}
To benchmark the performance of R2DL we compare it with the current established benchmark of a trained classifier using the same AMP training data set \cite{p_das}.

\begin{table}[h]
\caption{Baseline AMP Classification (Train from Scratch)}
\begin{tabular}{@{}lllllll@{}}
\toprule
\multicolumn{1}{c}{\multirow{2}{*}{Attribute}} & \multicolumn{3}{l}{Data-Split} & \multicolumn{2}{l}{Accuracy} & \multicolumn{1}{c}{\multirow{2}{*}{Screening Threshold}} \\ \cmidrule(lr){2-6}
\multicolumn{1}{c}{}        & Train & Valid & Test & Majority Class & Test & \multicolumn{1}{c}{} \\ \cmidrule(r){1-1} \cmidrule(l){7-7} 
\{Toxic, non-Toxic\}        & 8153  & 1019  & 1020 & 82.73          & 93.7 & -1.573 \\
\{AMP, non-AMP\}        & 6489  & 811  & 812 & 82.68.9          & 88.0 & 7.944 

              \\ \bottomrule
\end{tabular}
\end{table}

\subsection{Restricted Training Data Setting}
To further investigate the efficacy of the transfer learning approach, we compare the performance of R2DL versus the model trained from scratch with AMP data, with a restricted training data set. The test accuracies indicate that R2DL performs better when fewer labeled training data samples are available. Tables 2 and 3 show that when trained in a setting with fewer labeled data samples, R2DL outperforms the train from scratch method after the threshold of approximately 5000 samples. Below 5000 samples, both methods approximate random prediction, with R2DL not successfully transferring any learned representations \footnote{5000 AMP training samples and below were excluded from the results below as they showed statistically insignificant test accuracy.}.

\begin{table}[h]
\centering
\caption{Restricted Data Setting: Toxicity Prediction}
\label{tab:my-table}
\begin{tabular}{>{\centering\arraybackslash}m{5.0cm}|>{\centering\arraybackslash}m{2.5cm}|>{\centering\arraybackslash}m{2cm}|>{\centering\arraybackslash}m{2.5cm}>{\centering\arraybackslash}m{2cm}}

\toprule
Task            & AMP Sequences Training Samples & R2DL Test Accuracy & Bi-LSTM Test Accuracy (train from scratch)  \\ \midrule
Toxicity Prediction & 5000                 & 42.12              & 37.34                   \\
Toxicity Prediction & 6000                 & 62.98              & 49.62                     \\
Toxicity Prediction & 7000                 & 86.23              & 82.78                      \\
Toxicity Prediction & 8153                 & 89.34              & 93.7                     \\
 \bottomrule

\end{tabular}
\end{table}

\begin{table}[h]
\centering
\caption{Restricted Data Setting: AMP Prediction}
\label{tab:my-table}
\begin{tabular}{>{\centering\arraybackslash}m{5.0cm}|>{\centering\arraybackslash}m{2.5cm}|>{\centering\arraybackslash}m{2cm}|>{\centering\arraybackslash}m{2.5cm}>{\centering\arraybackslash}m{2cm}}

\toprule
Task            & AMP Sequences Training Samples & R2DL Test Accuracy & Bi-LSTM Test Accuracy (train from scratch)  \\ \midrule
AMP Prediction & 3500                 & 59.82              & 64.52                   \\
AMP Prediction & 4500                 & 72.76              & 68.41                     \\
AMP Prediction & 5500                 & 84.17            & 74.34                      \\
AMP Prediction & 6489                 & 90.01            & 88.0                    \\
 \bottomrule

\end{tabular}
\end{table}

\subsection{Repurposing Sentiment Classifiers}
BERT and GPT-3 are 2 of the most common powerful language models, that generalize well across various NLP tasks, dealing with language (sequence) data. In this experiment, we use BERT, a bidirectional transformer, tuned for the sentiment classification task on the IMDB movie review dataset. As R2DL assumed access to the gradients of the source model ($\nabla X$), this approach works in a semi-black box setting given that we access but do not modify the internal architecture. In this task, we use sentiment classification as the source task, in which there are 2 output classes (positive, negative), and AMP toxicity classification as the target task (toxic, non-toxic). The output-label mapping $h$ is then a simple 1-1 correspondence between (positive, toxic) and (negative, non-toxic). The input data of the source model (BERT) is tokenized on a word-level which form the atoms for our dictionary representation of $V_S$. The input data to the target task, AMP sequences, are tokenized on a character level with only 7 distinct tokens. The embeddings of a source vocabulary token, $v_s$, is then represented as a weighted combination of the atoms of the AMP sequence tokens. Using the $l^0$ norm in our objective function, 100 k-SVD iterations and $\epsilon = 0.045$, we are able to achieve accuracy on the order of the benchmark when trained from scratch in table 1. Table 4 shows accuracies for sentiment classification source models and the target task, and Table 5 shows accuracies as we increase the number of k-SVD iterations for a convergent solution.

\begin{table}[h]
\centering
\caption{AR for Sentiment Classifiers: Toxicity Prediction}
\label{tab:my-table}
\begin{tabular}{>{\centering\arraybackslash}m{5.0cm}|>{\centering\arraybackslash}m{1.7cm}|>{\centering\arraybackslash}m{1.7cm}|>{\centering\arraybackslash}m{1.7cm}|>{\centering\arraybackslash}m{1.7cm}}

\toprule
Source Model                     & AMP Sequence Samples & k-SVD Iterations & Training Accuracy & Test Accuracy \\ \midrule
BERT (Bidirectional Transformer) & 8153                 & 100              & 74.78             & 87.23         \\
BERT (Bidirectional Transformer) & 8153                 & 250              & 76.23             & 86.93         \\
Bi-LSTM                          & 8153                 & 100              & 68.34             & 81.25         \\ \bottomrule

\end{tabular}
\end{table}

\begin{table}[h]
\centering
\caption{k-SVD Iterations vs. Accuracy}
\label{tab:my-table}
\begin{tabular}{@{}lllll@{}}
\toprule
Source Model                                                                & AMP Sequence Samples & k-SVD Iterations & Training Accuracy & Test Accuracy \\ \midrule
BERT                                            & 8153                 & 100              & 74.78             & 87.23         \\
BERT                                            & 8153                 & 200              & 73.24             & 85.61         \\
BERT                                             & 8153                 & 300              & 75.12             & 87.89         \\ 
\bottomrule
\end{tabular}
\end{table}

From Table 5, we see that R2DL acheives the performance (89.34 \%) of the baseline (93.7.0 \%) when the pretrained classifier is a bidirectional transformer. Intuitively, we can understand why more expressive models have an increased representational capacity that can be leveraged to transfer relevant features from the pretrained classifier. Additionally, more precise k-SVD solutions beyond 100 iterations are not correlated with increased performance, and only increase computational cost, making this approach less efficient. 

\section{Conclusion}
This work formalizes the argument for finding sparse representations of dense input data for cross domain transfer learning. In a setting where access to well structured and labelled data is limited, we can leverage representations in deep models through adversarial reprogramming and approximating a sparse coding. This approach demonstrated a either higher or comparable performance across different tasks when compared to training from scratch, but at a much lower cost and also performed better on fewer target task training samples. Our results provide new insights for domain-agnostic transfer learning, and establishes avenues for several molecular learning tasks that have been constrained by a lack of access to well-defined datasets. 

\section{Future Work}
Immediate plans for this work include evaluating cost effectiveness of this approach with analysis of computational complexity and time. Future plans for this work extend to representation learning tasks for molecules with multi-dimensional structures (Simplified molecular-input line-entry system - SMILES). We also plan to explore meta-learning capabilities for generalization with respect to various molecular learning tasks, variable length molecule sequences, and variable sequence structures. Planned target tasks in the molecule domain include graph generation and genetic code classification.



\bibliographystyle{unsrt}

\bibliography{citations}

\end{document}